\documentclass{article}
\usepackage{acra_format/acra}
\usepackage[hidelinks]{hyperref}
\usepackage{graphics}
\usepackage[caption=false]{subfig}
\usepackage[font=small,labelfont=bf]{caption}% caption format
\usepackage{graphicx}
\usepackage{balance}

\graphicspath{{figures/}}% where to look for graphics
\usepackage{float}% for [H]
\usepackage{epsfig}% for postscript graphics files
\usepackage{svg}% i was trying to import svg diagrams
\usepackage{dsfont}% double struck
\usepackage{amsmath}% assumes amsmath package installed
\usepackage{amssymb}% assumes amsmath package installed
\usepackage{siunitx}% units
\usepackage{mathtools}% for shortintertext and other stuff
\usepackage{multirow}% table stuff
\usepackage{enumitem}% resuming lists
\newcommand{\PRM}[1]{PRM$^{#1}$}% PRM^* in text mode, to use type \PRM*

\newcommand{\numberthis}{\stepcounter{equation}\tag{\theequation}}
\DeclareMathOperator*{\argmin}{arg\,min}

\newcommand{\Expected}[1]{\mathds{E}\left[#1\right]}% Expected value
\DeclareMathOperator{\Var}{Var}
\newcommand{\Variance}[1]{\Var\left[#1\right]}% Variance
\newcommand{\norm}[1]{\left\lVert#1\right\rVert}
\newcommand{\magsq}[1]{\norm{#1}^2}% magnitude squared
\newcommand{\diff}[2][]{\mathop{\mathrm{d}^{#1}#2}}% upright 'd' at the end of integral
 \renewcommand{\vec}{\mathbf}% vectors as bold
\newcommand{\I}{\mathcal{I}}% invasiveness
\newcommand{\x}{\vec{x}}
\newcommand{\pos}{\x}
\newcommand{\vel}{\vec{v}}
\newcommand{\Vel}{\vec{V}\!}% This V is very wide
\newcommand{\DVel}{\Delta\!\Vel}
\newcommand{\meanVel}[1][]{\boldsymbol{\mu}_{\Vel#1}}
\newcommand{\varVel}[1][]{\sigma^2_{\Vel#1}}
\newcommand{\density}{\rho}
\renewcommand{\time}{t}% \time command gives minutes in day :shrug:
\newcommand{\Real}{\mathds{R}}
\newcommand{\PosSpace}{\mathds{X}}
\newcommand{\VelSpace}{\mathds{V}}
\newcommand{\robot}{r}
\newcommand{\ofrobot}{_\robot}
\newcommand{\atrobot}{_\robot}% not a property of the robot, but something 'at' the location of the robot

\usepackage{amsthm}
\newtheorem{problem}{Problem}

\theoremstyle{definition}

\title{Minimally Invasive Social Navigation}
\author{   
    Stefan~H.~Kiss, K.~Y.~Cadmus~To, Chanyeol Yoo, Robert Fitch and Alen Alempijevic\\
    University of Technology Sydney, NSW Australia \\
    \{stefan.h.kiss,cadmus.to\}@student.uts.edu.au,
    \{chanyeol.yoo,robert.fitch,alen.alempijevic\}@uts.edu.au
}

\begin{document}

\maketitle

\begin{abstract}
    %% SK should abstract follow these points, or more for introduction?
    % RF all scientific communication follows this pattern in one way or another. We have a nice
    % book that goes into this in detail. I think Brian has scanned excerpts on a server somewhere.
    Integrating mobile robots into human society involves the fundamental problem of navigation in crowds. This problem has been studied by considering the behaviour of humans at the level of individuals, but this representation limits the computational efficiency of motion planning algorithms. We explore the idea of representing a crowd as a flow field, and propose a formal definition of path quality based on the concept of invasiveness; a robot should attempt to navigate in a way that is minimally invasive to humans in its environment. We develop an algorithmic framework for path planning based on this definition and present experimental results that indicate its effectiveness. These results open new algorithmic questions motivated by the flow field representation of crowds and are a necessary step on the path to end-to-end implementations.
    % \begin{itemize}
    %     \item[Vision:]
    %         robotic integration into society
    %     \item[Motivation:]
    %         robots can do things humans and existing systems can't or don't want to do
    %     \item[Challenge:]
    %         robots can be dangerous,
    %         robots can interfere with the operation of existing systems
    %     \item[Approach:]
    %         \iitem formulate a measure of invasiveness
    %         \iitem explicitly minimise invasiveness during planning
    %     \item[Significance:]
    %         we can factor social costs into robot planning,
    %         robots can work alongside humans,
    %         yay
    % \end{itemize}
\end{abstract}

%%%%%%%%%%%%%%%%%%%%%%%%%%%%%%%%%%%%%%%%%%%%%%%%%%%%%%%%%%%%%%%%%%%%%%%%%%%%%%%%
%%%%%%%%%%%%%%%%%%%%%%%%%%%%%%%%%%%%%%%%%%%%%%%%%%%%%%%%%%%%%%%%%%%%%%%%%%%%%%%%
\section{Introduction}

% RF merged related work with intro, since it was written more as a nice intro than
% a stand-alone related work section
%%%%%%%%%%%%%%%%%%%%%%%%%%%%%%%%%%%%%%%%%%%%%%%%%%%%%%%%%%%%%%%%%%%%%%%%%%%%%%%%
%%%%%%%%%%%%%%%%%%%%%%%%%%%%%%%%%%%%%%%%%%%%%%%%%%%%%%%%%%%%%%%%%%%%%%%%%%%%%%%%
%\section{Related Work}

%1. We consider the problem of navigating a mobile robot through dense human crowds
Humans and robots co-existing in an environment require an understanding of each other's motion to perform safe interactions. For humans, the ability to predict the motion of others within an environment enables implicit path planning to reach intended goals in a self-centred manner. For robots, the capability to anticipate human motion can facilitate more fluent interaction~\cite{Hoffman2007}. 
We are interested in this fundamental problem in the case of dense human crowds, where a mobile robot must consider the motion of many humans in order to navigate. This problem is important to any application of mobile robots in crowded spaces, such as public indoor and outdoor areas. Our main focus is on how to use observed human motion to develop a path planning framework that seeks to find a path to a goal region while minimising interference with humans.

When individuals move in a crowd, the decision-making process is more than merely reactive behaviour~\cite{Trautman2015}; humans incorporate reasoning about the possible actions of others. Ignoring inter-dependencies in human motion may lead to overly conservative  motion planning~\cite{Turnwald2019}. This issue can be avoided if the robot can anticipate humans' cooperative collision avoidance and can take into account the goal-driven nature of human decision making. 

%2. Rather than using pair-wise interaction, we resort to dealing with flow
Encoding human cooperative interaction in motion planning can be coarsely divided into approaches that (1) build upon sets of social rules, or (2) consider reactive behaviours. Methods that encode social rules have evolved from Helbing's seminal work on social forces~\cite{Helbing1995social} to incorporate relative velocity and intended direction of travel~\cite{Moussaid2009}. Deep learning strategies have recently come into focus~\cite{Alahi2016}. Alternatively,~\cite{VanDenBerg2008} propose to take into account reactive behaviour, assuming humans can be modelled as agents that make similar collision-avoidance reasoning.

Integrating either of these approaches into a motion planning framework requires a means of evaluating pairwise interactions~\cite{Rudenko2018} and of inferring the goal of each agent in order to predict its future trajectory~\cite{Vemula2017,Bevilacqua2018}. The computation required to implement these operations depends on the number of pedestrians in the scene, and can become prohibitive in large crowd densities~\cite{Trautman2015}; the computational complexity of modelling pairwise interactions is linear in the number of pedestrians in the best case. Additionally, modelling accuracy is typically also dependent on time, contributing poor predictions at larger time scales.

%3. We differentiate our work from others by ...
Bootstrapping planning by exploiting social rules of spatial occupancy~\cite{Alempijevic2013} or relative velocity~\cite{Bera2019} has proved to be effective in socially-aware robot navigation in low to medium-density environments. However, to cater towards higher densities we argue that looking macroscopically at flow (both relative and average) will enable more scalable solutions to be developed. Recent analysis of human-robot interaction around emergency egress points indicates interplay between the direction of pedestrian flow~\cite{Chen2018} and the relative velocity between robot and pedestrian. In addition, the energy cost in the navigation process has been identified as being of vital importance~\cite{Chen2018}. We therefore posit that the objective function of a motion planner should be inherently goal driven (such as the goal-directed motion of humans) while social compliance should be related to minimal invasiveness~(minimal disruption to flow of others).

To operationalise these ideas, we encode human motion macroscopically as a flow field and propose a corresponding formal definition of invasiveness given the macroscopic properties of the crowd. This formulation is defined with respect to both deterministic and stochastic flows, and matches our intuition of invasiveness, which is a subjective concept that has no well-established mathematical representation. We then develop an algorithmic framework for non-myopic minimally-invasive social navigation using this definition. The non-myopic property, in this context, means that the algorithm aims to find a globally optimal solution as opposed to one that is locally optimal or reactive. 
Similar work has been done to navigate through wind fields or ocean currents~\cite{lee2017energy,yoo2016online,brian2019online,cadmus2019streamlines,Chanyeol2019,to2019streamlinebased}. To the best of our knowledge, our approach is the first to formulate social navigation problem using flow fields.
Our framework is presented using the well-known \emph{optimal Probabilistic RoadMap}~(\PRM*) algorithm for sampling-based motion planning, but other variants could be easily substituted.

We illustrate the behaviour of our algorithms in simulation of artificial scenarios that allow us to systematically examine a set of macroscopic~(high-level) features of crowd-based flow fields. Further, we consider an indoor scenario that shows how these features would appear in an application of a robot moving through a crowded building. Results show that the paths generated by our algorithm are less-invasive than those of comparison methods.

The main contributions of this paper are a novel formal definition of invasiveness given the macroscopic properties of a crowd, and a novel framework for non-myopic, minimally-invasive social navigation. The significance of this work is that it contributes a necessary first step towards the larger goal of finding computationally efficient solutions that are suitable for real-world applications. As a result of this work, a number of new problems arise that are motivated by the flow-field representation. We present an overview of these problems in the concluding section of the paper.

% \todo{include citations for~\cite{Treuille2006continuum} ``Continuum Crowds''}
%SK: don't see where to put it

%%%%%%%%%%%%%%%%%%%%%%%%%%%%%%%%%%%%%%%%%%%%%%%%%%%%%%%%%%%%%%%%%%%%%%%%%%%%%%%%
%%%%%%%%%%%%%%%%%%%%%%%%%%%%%%%%%%%%%%%%%%%%%%%%%%%%%%%%%%%%%%%%%%%%%%%%%%%%%%%%
\section{Problem Formulation}
The robot's environment consists of obstacles and a pedestrian crowd in 2D space~$\PosSpace \subset \Real^2$.
Our robot can manoeuvre through it independent of the crowd's behaviour.
Hence the first-order dynamics of this system is:
\begin{equation} \label{eq:dynamics}
    \dot\pos\ofrobot = \vel\ofrobot
    \,,
\end{equation}
where $\pos\ofrobot$ is the position of the robot in 2D space free from obstacles, and $\vel\ofrobot$ is its velocity in some limited velocity space~$\VelSpace \subseteq \Real^2$.
The trajectory of the robot over time~$\time$ is denoted as $\pos\ofrobot(\time)$.

The interaction of the robot's presence in the crowd is measured as \emph{social invasiveness}~$\I\ofrobot$.
This invasiveness arises from the instantaneous amount of influence of the robot amongst the pedestrians.
% This quantity typically corresponds to the sum of social forces~\cite{TODO} in the neighbourhood.
% This quantity is analogous to the sum of social force magnitudes~\cite{Helbing1995social,Moussaid2009} or of proximity-based discomfort values~\cite{Hoogendoorn2003simulation,Treuille2006continuum}.
This quantity is a function of robot and environment states.
Existing methods to measure invasiveness include social force~\cite{Helbing1995social,Moussaid2009} and proximity-based discomfort~\cite{Hoogendoorn2003simulation,Treuille2006continuum}.

In the interest of minimising the overall intrusion of the robot traversal through crowds, the path planning problem is formally defined as a minimisation of total interference along the path:
\begin{problem} [Minimally invasive path planning in crowds]
    Given a measure of invasiveness with the crowd~$\I\ofrobot$,
    % Given a representation of the crowd, a measure of invasiveness~$\I\ofrobot$,
    the robot's dynamics~\eqref{eq:dynamics}, its initial position~$\pos_\text{init}$, and goal position~$\pos_\text{goal}$, find the optimal trajectory
    \begin{align}
        \pos\ofrobot^*
            &=
                \argmin_{\pos\ofrobot}
                \int\limits_0^T
                    \I\ofrobot
                \diff\time
                \quad\text{for}\quad T\in\Real_{\geq0}
        \,,
        \label{eq:objective}
    \end{align}
    such that
    \begin{align*}
        \pos\ofrobot(0) &= \pos_\text{init}
        \,,\\
        \pos\ofrobot(T) &= \pos_\text{goal}
        \,.
    \end{align*}
\end{problem}

To address this problem we must first formulate an analytical measure of social invasiveness that alleviates the computational bottleneck during planning.
The optimum trajectory is then found through the application of sampling-based motion planning minimising the total invasiveness along the trajectory.

%%%%%%%%%%%%%%%%%%%%%%%%%%%%%%%%%%%%%%%%%%%%%%%%%%%%%%%%%%%%%%%%%%%%%%%%%%%%%%%%
%%%%%%%%%%%%%%%%%%%%%%%%%%%%%%%%%%%%%%%%%%%%%%%%%%%%%%%%%%%%%%%%%%%%%%%%%%%%%%%%
\section{Social Invasiveness}
Measures of social invasiveness can be formulated by treating pedestrians as point masses, however the minimum time complexity of these approaches are linear.
This implies that their computation becomes prohibitively expensive in dense crowds.
Additionally, modelling accuracy is typically also dependent on the prediction horizon time, contributing poor predictions at larger time scales.

However, we propose that there exists properties of pedestrian crowds that remain fairly consistent over time \cite{Treuille2006continuum,Karamouzas2014universal} which describe their macroscopic behaviour.
    % , and can be used to describe a definition of social invasiveness
We present our formulation of social invasiveness based on these ideas.

% Stefan wrote this bit which might fit here >>>
% Previous approaches have tried to model the behaviour of pedestrians as interacting point masses, with some success.
% However, such approaches have severe limitations.

% Firstly, the computational complexity/cost of modelling the pairwise interactions between pedestrians scales with the number of pedestrians, linearly in the best case.
%     \todo{Na\"ive~$O(n^2)$, k-d tree~$O(n \log n)$, Alex's~$O(n)$.}
%     In the case of planning through such models, such modelling also scales linearly with
%         % time.
%         time horizon,
%         the time taken for the robot to reach the goal.

% Additionally, modelling accuracy is typically also dependent on the time, contributing poor predictions at larger time scales.

% For large and dense crowds, and planning over longer time horizons, such computation becomes prohibitively expensive, and the produced models inaccurate.

% To improve/fix/deal with, we notice that, at larger scales, the properties of pedestrian crowds remain fairly consistent over time \cite{Karamouzas2014universal,?}.
% As such, we find a definition/model of the expected \emph{social invasiveness} in terms of macroscopic properties of the pedestrian crowd in consideration.
%     Thus, our ``modelling'' has constant computational complexity \wrt the number of pedestrians and plan duration. ($O(1)$ vs at least $O(nT)$)
% <<<

%%%%%%%%%%%%%%%%%%%%%%%%%%%%%%%%%%%%%%%%%%%%%%%%%%%%%%%%%%%%%%%%%%%%%%%%%%%%%%%%
\subsection{Invasiveness in Crowd Flows}
Before we present our analytical definition of social invasiveness, we begin by introducing the continuous representation of crowds.
% As crowd behaviours approach equilibrium, they can be modelled as time-invariant flow fields with density.
Consistent crowd behaviours can be modelled as time-invariant flow fields with density.
A \emph{crowd flow} is defined as the 2D density and velocity field pair~$(\density,\Vel)$, where
\begin{align*}
    \density&: \PosSpace \to \Real_{\geq 0}
    \,,\\
    \Vel&: \PosSpace \to \Real^2
    \,.
\end{align*}
The density of the crowd~$\density(\pos)$ is a scalar field that describes the expected number of pedestrians per unit of area at~$\pos$.
The crowd velocity~$\Vel(\pos)$ is a vector field corresponding to the speed and direction of the pedestrians at $\pos$.

We can now define the invasiveness of the robot as
    \begin{align}
        \I\ofrobot &= \density\atrobot \magsq{\DVel\atrobot}
        \label{eq:invasiveness}
        \,,
    \end{align}
    where
    \begin{align*}
        \density\atrobot &= \density(\pos\ofrobot)
        \,,\\
        \Vel\atrobot &= \Vel(\pos\ofrobot)
        \,\text{, and}\\
        \DVel\atrobot &= \Vel\atrobot - \vel\ofrobot
        \,.
    \end{align*}

This formulation consists of two factors which account for different aspects of invasiveness.
The first factor~$\rho \norm\DVel$ is proportional to the expected number of pedestrians per second that cross a \emph{region of interaction} around the robot.
    The constant of proportionality can be thought of as related to the cross-sectional width of this region.
A second factor~$\norm\DVel$ assumes that the interference caused upon interaction is proportional to the velocity difference between the two parties, and is therefore inversely proportional to the available time to act.
% I think this should be the only occurance of "interference"

Equation~\eqref{eq:invasiveness} matches our intuition of what we expect to be invasive.
The robot will be less invasive if it travels through areas of low crowd density or if it travels at the same velocity as the crowd.

Crowd flows start to become poor models of pedestrian behaviour as multiple flows overlap~\cite{Jodoin2013metatracking}.
The velocity at each point can no longer described by a single vector as pedestrian intents are not dictated solely by their position in space.

%%%%%%%%%%%%%%%%%%%%%%%%%%%%%%%%%%%%%%%%%%%%%%%%%%%%%%%%%%%%%%%%%%%%%%%%%%%%%%%%
\subsection{Invasiveness in Stochastic Crowd Flows}
To account for multiple overlapping pedestrian flows, we consider \emph{stochastic crowd flows}.
More specifically, we augment the concept of crowd flows with a variance of velocity.
This allows us to rigorously reinterpret the crowd flow in a probabilistic sense,
    and redefine social invasiveness in expectation:
\begin{align*}
    \I\ofrobot
        &=
            \Expected{
                \density\atrobot
                \magsq{\DVel\atrobot}
            }
        \,,
\end{align*}
from which we can derive
\begin{align}
    \I\ofrobot
        &=
            \density\atrobot
            \left(
                \magsq{
                    \meanVel[\atrobot]
                    -
                    \vel\ofrobot
                }
                +
                \varVel[\atrobot]
            \right)
    \,,
    \label{eq:expected_invasiveness}
\end{align}
where
\begin{gather*}
\begin{aligned}
    \meanVel &: \PosSpace \to \Real^2
    \,,&
    \meanVel &: \pos \mapsto \Expected{\Vel(\pos)}
    \,,\\
    \varVel &: \PosSpace \to \Real_{\geq0}
    \,,&
    \varVel &: \pos \mapsto \Variance{\Vel(\pos)}
    \,,
\end{aligned}
\label{eq:velfields}
\\
\begin{aligned}
    \meanVel[\atrobot] &= \meanVel(\pos\ofrobot)
    \,\text{, and}\\
    \varVel[\atrobot] &= \varVel(\pos\ofrobot)
    \,.
\end{aligned}
\end{gather*}
% \begin{align}
%     \meanVel[\atrobot] &= \meanVel(\pos\ofrobot) = \Expected{\Vel(\pos\ofrobot)}
%     \,\text{, and}\\
%     \varVel[\atrobot] &= \varVel(\pos\ofrobot) = \Variance{\Vel(\pos\ofrobot)}
%     \,.
% \end{align}
Note that $\Variance{\cdot}$ here indicates the \emph{scalar} variance, defined as the expected value of the squared euclidean distance from the mean:
% , or equivalently as the trace of the covariance matrix:
    $\Variance{\x}
        = \Expected{\magsq{\pos - \Expected{\pos}}}
        % = \text{tr}(\text{Cov}(\x,\x))
    $.

These additional macroscopic properties of the crowd can be understood intuitively.
The mean velocity~$\meanVel(\pos)$ describes the typical flow of pedestrians around point~$\pos$.
The variance of velocity~$\varVel(\pos)$ describes the irregularity of the flow around that point.
A low variance indicates a coherent flow of pedestrians in the mean direction, whereas a large velocity variance with a mean close to zero indicates that pedestrians walk through the region from multiple directions with no dominant flow.
% In practice, variance can never be zero due to errors from the model of the crowd flow.
In practice, variance is always expected to be greater than zero as crowd flows approximate multiple individuals with different destinations.

%%%%%%%%%%%%%%%%%%%%%%%%%%%%%%%%%%%%%%%%%%%%%%%%%%%%%%%%%%%%%%%%%%%%%%%%%%%%%%%%
%%%%%%%%%%%%%%%%%%%%%%%%%%%%%%%%%%%%%%%%%%%%%%%%%%%%%%%%%%%%%%%%%%%%%%%%%%%%%%%%
\section{Minimally Invasive Path Planning in Crowd Flows}
To find a minimally invasive trajectory with general path planning algorithms, we can express invasiveness between two points as a cost function.
Assuming that the robot travels in a straight line between points, we need to determine its speed along the path.

%%%%%%%%%%%%%%%%%%%%%%%%%%%%%%%%%%%%%%%%%%%%%%%%%%%%%%%%%%%%%%%%%%%%%%%%%%%%%%%%
\subsection{Minimally Invasive Speed}
As the robot travels along a straight line segment, the invasiveness along an infinitesimal step of length~$ds$ can be expressed as
\begin{align}
    \I\ofrobot d\time = \I\ofrobot \frac{ds}{v\ofrobot} 
    \,,
\end{align}
where $v\ofrobot$ is the robot's speed towards the goal.
We can now determine the minimally invasive speed for the infinitesimal step:
\begin{align*}
    v\ofrobot^*
        &=
            \argmin_{v\ofrobot} \frac{\I\ofrobot}{v\ofrobot}
        \\
    v\ofrobot^*
        &=
            \sqrt{
                \magsq{\meanVel[\atrobot]}
                + \varVel[\atrobot]
            }
    \numberthis
    \label{eq:optimal_vel}
\end{align*}
Interestingly, this result implies that the optimal instantaneous speed does not depend on the direction of motion itself.

% Note that this velocity~$v\ofrobot^*\hat\vel\ofrobot$ may not be within the valid set~$\VelSpace$, however, it is sufficient to select the valid speed~$v\ofrobot$ closest to~$v\ofrobot^*$ as this optimisation is convex with one minimum.
%SK: This is wrong. Well, the idea is right, but the wording is wrong, I tried to generalise too far.

This analytical value of minimally invasive speed allows us to calculate the invasiveness along a path by integrating with the appropriate discretisation.

%%%%%%%%%%%%%%%%%%%%%%%%%%%%%%%%%%%%%%%%%%%%%%%%%%%%%%%%%%%%%%%%%%%%%%%%%%%%%%%%
% \subsection{\PRM* Integration}
\subsection{Integration with Optimal Path Planner}
We use \PRM* to generate a bidirectional graph using samples randomly drawn from a bounded 2D configuration space of position, including the initial and goal positions.
%SK: 'bidirectional' is maybe confusing, but it is a 'symmetric directed' graph.
The weights of the graph edges correspond to the invasiveness between two nodes.
In this context, the straight line assumption for invasiveness along these edges becomes negligible as distance between considered samples approaches zero as the number of sample points increases~\cite{Karaman2007sampling}.
%SK: Wtf is this sentence saying?
%SK: Saying that as the number of samples increases, solutions approach continuous (non-piecewise-linear) trajectories

Dijkstra's algorithm is then applied on the resulting graph with respect to the starting position to find a minimally invasive tree to every other node in the graph.
Trajectories generated this way simultaneously avoid obstacles while approaching the optimal trajectory as the number of samples increases.
% \todo{Some talk about how global planning allows for non-myopic solutions.}

% To perform the trajectory optimisation of equation \eqref{eq:objective}, we employ the probabilistically-complete and asymptotically-optimal sampling based planner \PRM* \cite{Karaman2007sampling}.
% However, \todo{others planners would work too.}

% \todo{This means our approach gains all the advantages of sampling based planners, such as dealing with obstacles,\dots}

% The edge connections are integrated over at a sufficiently fine spatial resolution to calculate the total interference and duration (as velocity is calculated analytically).

% Finally, Dijkstra's algorithm is used to find the minimally invasive route through the graph.

% \todo{Some talk about how generating a roadmap provides a query-able structure appropriate for the online control of robot also locally avoiding pedestrians?}

%%%%%%%%%%%%%%%%%%%%%%%%%%%%%%%%%%%%%%%%%%%%%%%%%%%%%%%%%%%%%%%%%%%%%%%%%%%%%%%%
%%%%%%%%%%%%%%%%%%%%%%%%%%%%%%%%%%%%%%%%%%%%%%%%%%%%%%%%%%%%%%%%%%%%%%%%%%%%%%%%
\section{Experiments}
We test our algorithm on different simple scenarios demonstrating its behaviour when one macroscopic property is varied and a complex scenario where a combination of them is varied.
The macroscopic parameters we consider is the crowd density, velocity, and the variance of the velocities.
The path that our algorithm generates is compared against a planner that only aims to reduce path length and ignores the crowd flow.
Understandably, the trajectories generated from our planner have lower invasiveness than the baseline in all cases.
% \todo{Can mention any other parameters that are consistent between simple and complex cases here}

%%%%%%%%%%%%%%%%%%%%%%%%%%%%%%%%%%%%%%%%%%%%%%%%%%%%%%%%%%%%%%%%%%%%%%%%%%%%%%%%
\subsection{Independent Comparisons}

\begin{table}[!tb]
    \caption{Parameters and results for comparing macroscopic properties}
    \tiny %AA - added to fit table into column
    \begin{tabular}{rccccc}
    \multirow{3}{*}{Scenario} & \multicolumn{3}{c}{\shortstack{macroscopic\\properties}} & \multicolumn{2}{c}{\shortstack{interference\\of trajectory}} \\
                  &$\density  $&$\meanVel      $&$\varVel    $& social & na\"ive \\
                  &$\si{1/m^2}$&$\si{m/s}      $&$\si{m^2/s^2}$\\\hline
    density       &$ 0.5-1.5  $&$\vec0         $&$     1     $&$  17.1   $&$  27.7  $\\
    velocity      &$    1     $&$\norm{\cdot}=1$&$    0.25   $&$   5.5   $&$  22.3  $\\
    variance      &$    1     $&$\vec0         $&$ 0.25-1.25 $&$  23.1   $&$  24.5  $
    \end{tabular}
    \label{tab:simpleParameters}
\end{table}

Each of the macroscopic properties, density, velocity and variance, are varied independently while others stay uniform across the space.
Table~\ref{tab:simpleParameters} shows the 3 different scenarios named after the parameter that varies in them.

\begin{figure}[!tb]
    \centering
    \subfloat[Density heatmap]{
        \includegraphics[height=0.62\columnwidth]{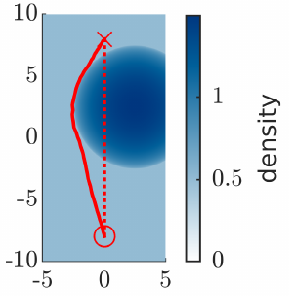}
        \label{subfig:density_env}
    }%
    \subfloat[Invasiveness edges]{
        \includegraphics[height=0.62\columnwidth]{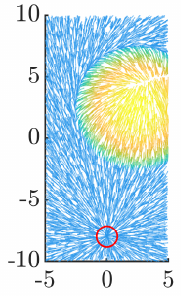}
        \label{subfig:density_tree}
    }
    \caption{
        Simple case varying crowd density.
        % The trajectory from our social planner is shown as a solid line; the naive planner shown as a dotted line.
        The left compares the trajectory from the na\"ive planner and our social planner, shown as dotted and solid lines respectively, from an initial position (circle) to a goal (cross).
        \PRM*'s minimal invasive tree from the initial position is shown on the right, with edges coloured by the invasiveness per unit distance.
    }
    \label{fig:density}
\end{figure}

The density scenario is analogous to a stationary crowd gathered around a point of interest.
Figure~\ref{subfig:density_env} shows our algorithm avoiding the dense crowd, as it is more invasive to wade through the crowd than it is to move around it through less dense regions.
In the less dense regions, the algorithm prefers straight paths to minimise the time spent around the pedestrians.% despite fewer people.

It is interesting to see critical points in the minimally invasive tree in figure~\ref{subfig:density_tree} in the upper right of the crowd which we will refer to as \emph{breakpoints}.
These breakpoints indicate regions where small changes in goal specification drastically changes the optimal path.
In this case, the breakpoint is not centred at the dense crowd region, but is more towards the far side of the crowd.
This is intuitive since if we need to plan a path to the middle of the crowd, we can only minimise the invasiveness of a path to the boundary of this crowd, which corresponds to a straight line in this scenario.

\begin{figure}[!tb]
    \centering
    % NOTE: We use height=* here because the figures have equal height
    \subfloat[Velocity vector field]{
        \includegraphics[height=0.68\columnwidth]{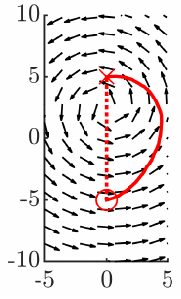}
        \label{subfig:velocity_env}
    }\hfill
    \subfloat[Invasiveness edges]{
        \includegraphics[height=0.68\columnwidth]{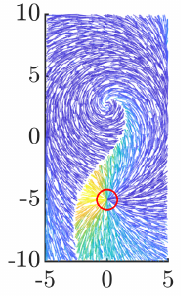}
        \label{subfig:velocity_tree}
    }
    \caption{
        Simple case varying crowd velocity.
        % The trajectory from our social planner is shown as a solid line; the naive planner shown as a dotted line.
        The left compares the trajectory from the na\"ive planner and our social planner, shown as dotted and solid lines respectively, from an initial position (circle) to a goal (cross).
        \PRM*'s minimal invasive tree from the initial position is shown on the right, with edges coloured by the invasiveness per unit distance.
    }
    \label{fig:velocity}
\end{figure}

The abstract scenario varying velocity corresponds to a crowd of people orbiting a point.
In this situation, we can travel with minimal invasiveness by following the crowd's movement.
By specifying a particular destination, our algorithm generates an interesting trajectory illustrated in figure~\ref{subfig:velocity_env}.
Following our intuition, it travels in the direction of flow.
On close inspection we also note that the trajectory bulges over time.
Since the destination is not directly ``downstream'' of the starting location, the robot must cross the flow of the crowd.
The rate at which the robot crosses the flow appears consistent through the trajectory as well since this would minimise the sum square of relative velocities.
This effect can be seen across the minimal invasive tree in figure~\ref{subfig:velocity_tree}.
The breakpoints in this scenario highlight the regions where it is equally invasive to move a short distance against the flow as to move a much longer distance with the flow.

\begin{figure}[!tb]
    \centering
    \subfloat[Variance heatmap]{
        \includegraphics[height=0.62\columnwidth]{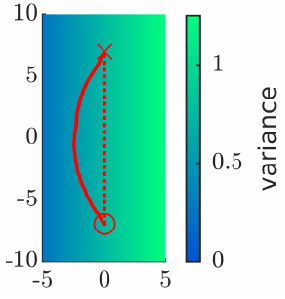}
        \label{subfig:variance_env}
    }%
    \subfloat[Invasiveness edges]{
        \includegraphics[height=0.62\columnwidth]{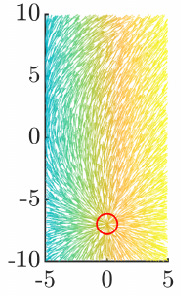}
        \label{subfig:variance_tree}
    }
    \caption{
        Simple case varying the variance of the crowd velocity.
        % The trajectory from our social planner is shown as a solid line; the naive planner shown as a dotted line.
        The left compares the trajectory from the na\"ive planner and our social planner, shown as dotted and solid lines respectively, from an initial position (circle) to a goal (cross).
        \PRM*'s minimal invasive tree from the initial position is shown on the right, with edges coloured by the invasiveness per unit distance.
    }
    \label{fig:variance}
\end{figure}

The final simple scenario varies the variance of crowd velocities, as a result of summing 3 pedestrian flows.
Two flows travel in opposite directions in the vertical axis in figure~\ref{subfig:variance_env} with equal speed, but with more pedestrians on the right side.
A third flow is added with zero velocity to ensure that the density of the crowd across the space is even.

In this case, our planner finds a trajectory that is the compromise between travelling longer in regions with lower variance and travelling shorter in regions with higher variance.
It is intuitive to avoid regions towards the right of the environment despite some of the crowd going in the robot's direction since there are also others with the opposite velocity travelling against the robot.
In practice, humans avoid causing regions of high variance of velocity as they are associated with higher rates of collision which leads to social phenomena such as lane forming.%~\cite{TODO}.

%%%%%%%%%%%%%%%%%%%%%%%%%%%%%%%%%%%%%%%%%%%%%%%%%%%%%%%%%%%%%%%%%%%%%%%%%%%%%%%%
\subsection{Concert Hall}

\begin{figure}[!tb]
	\centering
	\subfloat[Density heatmap and velocity vector field] {
	    \includegraphics[width=0.9\columnwidth]{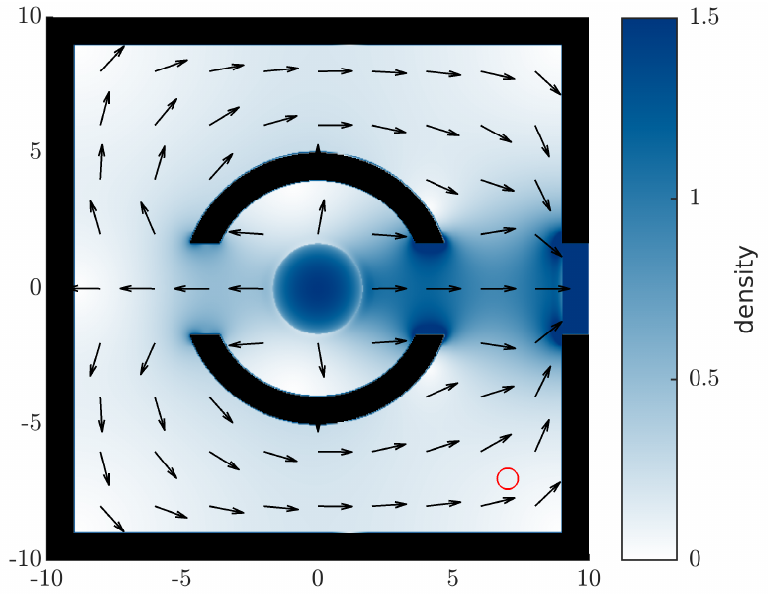}
	    \label{subfig:hall_env}
    }
    \\
	\subfloat[Invasiveness edges] {
	    \includegraphics[width=0.9\columnwidth]{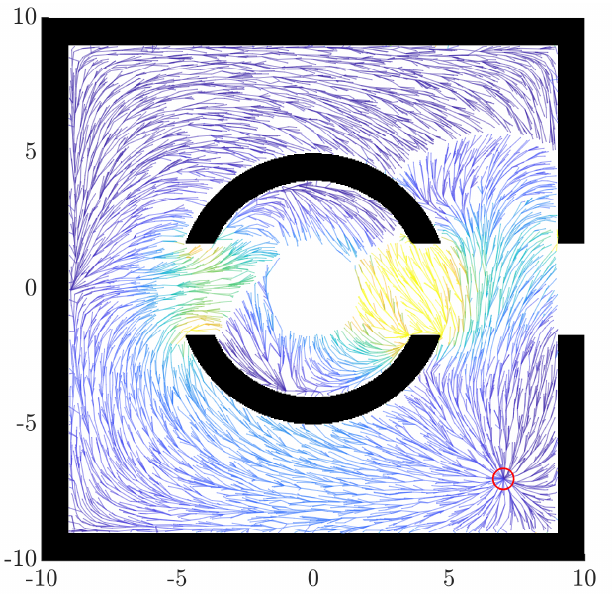} \label{subfig:hall_tree}
    }
    \caption{
        A hypothetical concert hall scene, showing the application of our work to a larger and more complex environment.
        \PRM*'s minimal invasive tree from an initial position (circle) is shown below, with edges coloured by the invasiveness per unit distance.
    }
	\label{fig:hall}
\end{figure}

We construct a hypothetical scene corresponding to audience members vacating a concert hall shown in figure~\ref{subfig:hall_env}.
The audience leaves the inner room through the two exits and then leaves the outer room through one main exit.
% More people form around the logical right inner door and the outer door.
Logically, more people form around the right inner door and the outer door.

% We visualise minimally invasive trajectories from all positions in the hall to the exit in figure~\ref{subfig:hall_tree}.
% Once again we see breakpoints that are away from the exit due to the preference to move along with the crowd to the exit.
In figure~\ref{subfig:hall_tree} we visualise the minimally invasive trajectories from an initial position in the bottom-right
to every position in the hall.
As the colour of each edge indicates the invasiveness per unit of distance, it can be seen that it is very costly to enter the inner room through the right door, or to cross between the the inner and outer doors.
Additionally, the breakpoint in the top right indicates the region where it is less invasive to move around the inner room than to push through the dominant crowd flow towards the main exit.

% This example highlights the algorithm's capabilities to handle varying crowds with varying velocities and densities.
This example illustrates that the algorithm is capable of planning over long time horizon to avoid myopic decisions in the presence of non-uniform velocity and density fields.

%%%%%%%%%%%%%%%%%%%%%%%%%%%%%%%%%%%%%%%%%%%%%%%%%%%%%%%%%%%%%%%%%%%%%%%%%%%%%%%%
%%%%%%%%%%%%%%%%%%%%%%%%%%%%%%%%%%%%%%%%%%%%%%%%%%%%%%%%%%%%%%%%%%%%%%%%%%%%%%%%
\section{Discussion and Future Work}
We have presented a new perspective on the problem of mobile robot navigation in dense crowds that is based on a flow field representation. This representation is advantageous in that it enables the development of computationally efficient planning algorithms whose running time is independent of the total number of individual pedestrians. Our results provide initial evidence to support this idea, but there are a number of areas of future work that would need to be addressed in order to develop an end-to-end implemented system.

One important question is how to create a flow field estimate given observations that could be readily acquired by a perception system. One approach would be to adapt recent results for estimating incompressible flows using specialised Gaussian process regression \cite{brian2019online}. Another important avenue to explore is the case of dynamic (time-dependent) flows. This is an instance of a time-dependent shortest path problem, which remains open in general. However, promising efficient solutions have recently been developed for relevant special cases including in our previous work \cite{James2019}. Finally, an interesting question for real-world systems is how to validate the fidelity of a flow field representation of crowds. This could be approached by comparing predicted behaviour of individuals, drawn from a distribution induced by the flow field, with observations of individuals in actual crowds.
%SK: 'including in our previous work' doesn't sound right to me
%SK: also the last sentence 'Finally ...' is partially linked to point 1 and not point 2, and I don't really think is able to stand alone
%SK: also, maybe some newlines in here?

% \begin{itemize}
%     \item Can use Brian's GP to capture mass flow rate as it is incompressible 
%     \item Can use James' Time dependant thing \cite{James2019}
% \end{itemize}

%%%%%%%%%%%%%%%%%%%%%%%%%%%%%%%%%%%%%%%%%%%%%%%%%%%%%%%%%%%%%%%%%%%%%%%%%%%%%%%%
%%%%%%%%%%%%%%%%%%%%%%%%%%%%%%%%%%%%%%%%%%%%%%%%%%%%%%%%%%%%%%%%%%%%%%%%%%%%%%%%
\section*{Acknowledgments}
This work is supported in part by Australian Government Research Training Program~(RTP) Scholarships and the University of Technology Sydney.

%%%%%%%%%%%%%%%%%%%%%%%%%%%%%%%%%%%%%%%%%%%%%%%%%%%%%%%%%%%%%%%%%%%%%%%%%%%%%%%%
%%%%%%%%%%%%%%%%%%%%%%%%%%%%%%%%%%%%%%%%%%%%%%%%%%%%%%%%%%%%%%%%%%%%%%%%%%%%%%%%
%\balance
\bibliographystyle{acra_format/named}
\bibliography{references}

\begin{thebibliography}{}

\bibitem[\protect\citeauthoryear{{Alahi} \bgroup \em et al.\egroup
  }{2016}]{Alahi2016}
A.~{Alahi}, K.~{Goel}, V.~{Ramanathan}, A.~{Robicquet}, L.~{Fei-Fei}, and
  S.~{Savarese}.
\newblock {Social LSTM: human trajectory prediction in crowded spaces}.
\newblock In {\em Proc. of IEEE Conference on Computer Vision and Pattern
  Recognition}, pages 961--971, 2016.

\bibitem[\protect\citeauthoryear{{Alempijevic} \bgroup \em et al.\egroup
  }{2013}]{Alempijevic2013}
A.~{Alempijevic}, R.~{Fitch}, and N.~{Kirchner}.
\newblock Bootstrapping navigation and path planning using human positional
  traces.
\newblock In {\em Proc. of IEEE ICRA}, pages 1242--1247, 2013.

\bibitem[\protect\citeauthoryear{Bera \bgroup \em et al.\egroup
  }{2019}]{Bera2019}
Aniket Bera, Tanmay Randhavane, and Dinesh Manocha.
\newblock The emotionally intelligent robot: Improving socially-aware human
  prediction in crowded environments.
\newblock In {\em Proc. of IEEE Conference on Computer Vision and Pattern
  Recognition Workshops}, June 2019.

\bibitem[\protect\citeauthoryear{{Bevilacqua} \bgroup \em et al.\egroup
  }{2018}]{Bevilacqua2018}
P.~{Bevilacqua}, M.~{Frego}, D.~{Fontanelli}, and L.~{Palopoli}.
\newblock Reactive planning for assistive robots.
\newblock {\em IEEE Robotics and Automation Letters}, 3(2):1276--1283, 2018.

\bibitem[\protect\citeauthoryear{{Chen} \bgroup \em et al.\egroup
  }{2018}]{Chen2018}
Z.~{Chen}, C.~{Jiang}, and Y.~{Guo}.
\newblock Pedestrian-robot interaction experiments in an exit corridor.
\newblock In {\em Proc. of International Conference on Ubiquitous Robots},
  pages 29--34, 2018.

\bibitem[\protect\citeauthoryear{Helbing and
  Moln{\'{a}}r}{1995}]{Helbing1995social}
Dirk Helbing and P{\'{e}}ter Moln{\'{a}}r.
\newblock {Social force model for pedestrian dynamics}.
\newblock {\em Physical Review E}, 51:4282--4286, 1995.

\bibitem[\protect\citeauthoryear{Hoffman and Breazeal}{2007}]{Hoffman2007}
Guy Hoffman and Cynthia Breazeal.
\newblock Effects of anticipatory action on human-robot teamwork efficiency,
  fluency, and perception of team.
\newblock In {\em Proc. of ACM/IEEE International Conference on Human-Robot
  Interaction}, pages 1--8, 2007.

\bibitem[\protect\citeauthoryear{Hoogendoorn and
  Bovy}{2003}]{Hoogendoorn2003simulation}
Serge Hoogendoorn and Piet~H.L. Bovy.
\newblock {Simulation of pedestrian flows by optimal control and differential
  games}.
\newblock {\em Optimal Control Applications and Methods}, 24(3):153--172, 2003.

\bibitem[\protect\citeauthoryear{Jodoin \bgroup \em et al.\egroup
  }{2013}]{Jodoin2013metatracking}
Pierre-Marc Jodoin, Yannick Benezeth, and Yi~Wang.
\newblock Meta-tracking for video scene understanding.
\newblock In {\em 2013 10th IEEE International Conference on Advanced Video and
  Signal Based Surveillance}, pages 1--6. IEEE, 2013.

\bibitem[\protect\citeauthoryear{Karaman and
  Frazzoli}{2011}]{Karaman2007sampling}
Sertac Karaman and Emilio Frazzoli.
\newblock {Sampling-based algorithms for optimal motion planning}.
\newblock {\em The International Journal of Robotics Research}, 30(7):846--894,
  2011.

\bibitem[\protect\citeauthoryear{Karamouzas \bgroup \em et al.\egroup
  }{2014}]{Karamouzas2014universal}
Ioannis Karamouzas, Brian Skinner, and Stephen~J Guy.
\newblock {Universal power law governing pedestrian interactions}.
\newblock {\em Physical Review Letters}, 113(23):238701:1--238701:5, 2014.

\bibitem[\protect\citeauthoryear{Lee \bgroup \em et al.\egroup
  }{2017}]{lee2017energy}
James Ju~Heon Lee, Chanyeol Yoo, Raewyn Hall, Stuart Anstee, and Robert Fitch.
\newblock Energy-optimal kinodynamic planning for underwater gliders in flow
  fields.
\newblock In {\em Proc. of ARAA ACRA}, 2017.

\bibitem[\protect\citeauthoryear{Lee \bgroup \em et al.\egroup
  }{2019a}]{James2019}
James Ju~Heon Lee, Chanyeol Yoo, Stuart Anstee, and Robert Fitch.
\newblock {Efficient optimal planning in non-FIFO time-dependent flow fields}.
\newblock In {\em arXiv:1909.02198 [cs.RO]}, pages 1--10, 2019.

\bibitem[\protect\citeauthoryear{Lee \bgroup \em et al.\egroup
  }{2019b}]{brian2019online}
Ki~Myung~Brian Lee, Chanyeol Yoo, Ben Hollings, Stuart Anstee, Shoudong Huang,
  and Robert Fitch.
\newblock {Online estimation of ocean current from sparse GPS data for
  underwater vehicles}.
\newblock In {\em Proc. of IEEE ICRA}, pages 3443--3449, 2019.

\bibitem[\protect\citeauthoryear{Moussa{\"\i}d \bgroup \em et al.\egroup
  }{2009}]{Moussaid2009}
Mehdi Moussa{\"\i}d, Dirk Helbing, Simon Garnier, Anders Johansson, Maud Combe,
  and Guy Theraulaz.
\newblock Experimental study of the behavioural mechanisms underlying
  self-organization in human crowds.
\newblock {\em Proceedings of the Royal Society B: Biological Sciences},
  276(1668):2755--2762, 2009.

\bibitem[\protect\citeauthoryear{{Rudenko} \bgroup \em et al.\egroup
  }{2018}]{Rudenko2018}
A.~{Rudenko}, L.~{Palmieri}, and K.~O. {Arras}.
\newblock Joint long-term prediction of human motion using a planning-based
  social force approach.
\newblock In {\em Proc. of IEEE ICRA}, pages 4571--4577, 2018.

\bibitem[\protect\citeauthoryear{To \bgroup \em et al.\egroup
  }{2019a}]{to2019streamlinebased}
Kwun Yiu~Cadmus To, James Ju~Heon Lee, Chanyeol Yoo, Stuart Anstee, and Robert
  Fitch.
\newblock Streamline-based control of underwater gliders in 3d environments.
\newblock In {\em Proc. of IEEE CDC}, 2019.

\bibitem[\protect\citeauthoryear{To \bgroup \em et al.\egroup
  }{2019b}]{cadmus2019streamlines}
Kwun Yiu~Cadmus To, Ki~Myung Brian~Lee Lee, Chanyeol Yoo, Stuart Anstee, and
  Robert Fitch.
\newblock Streamlines for motion planning in underwater currents.
\newblock {\em Proc. of IEEE ICRA}, 2019.

\bibitem[\protect\citeauthoryear{Trautman \bgroup \em et al.\egroup
  }{2015}]{Trautman2015}
Pete Trautman, Jeremy Ma, Richard~M Murray, and Andreas Krause.
\newblock Robot navigation in dense human crowds: Statistical models and
  experimental studies of human--robot cooperation.
\newblock {\em The International Journal of Robotics Research}, 34(3):335--356,
  2015.

\bibitem[\protect\citeauthoryear{Treuille \bgroup \em et al.\egroup
  }{2006}]{Treuille2006continuum}
Adrien Treuille, Seth Cooper, and Zoran Popovi{\'c}.
\newblock Continuum crowds.
\newblock In {\em ACM Transactions on Graphics (TOG)}, volume~25, pages
  1160--1168. ACM, 2006.

\bibitem[\protect\citeauthoryear{Turnwald and Wollherr}{2019}]{Turnwald2019}
Annemarie Turnwald and Dirk Wollherr.
\newblock Human-like motion planning based on game theoretic decision making.
\newblock {\em International Journal of Social Robotics}, 11(1):151--170, 2019.

\bibitem[\protect\citeauthoryear{{Van den Berg} \bgroup \em et al.\egroup
  }{2008}]{VanDenBerg2008}
J.~{Van den Berg}, {Ming Lin}, and D.~{Manocha}.
\newblock Reciprocal velocity obstacles for real-time multi-agent navigation.
\newblock In {\em Proc. of IEEE ICRA}, pages 1928--1935, 2008.

\bibitem[\protect\citeauthoryear{{Vemula} \bgroup \em et al.\egroup
  }{2017}]{Vemula2017}
A.~{Vemula}, K.~{Muelling}, and J.~{Oh}.
\newblock Modeling cooperative navigation in dense human crowds.
\newblock In {\em Proc. of IEEE ICRA}, pages 1685--1692, 2017.

\bibitem[\protect\citeauthoryear{Yoo \bgroup \em et al.\egroup
  }{2016}]{yoo2016online}
Chanyeol Yoo, Robert Fitch, and Salah Sukkarieh.
\newblock Online task planning and control for fuel-constrained aerial robots
  in wind fields.
\newblock {\em Int. J. Robot. Res.}, 35(5):438--453, 2016.

\bibitem[\protect\citeauthoryear{Yoo \bgroup \em et al.\egroup
  }{2019}]{Chanyeol2019}
Chanyeol Yoo, Stuart Anstee, and Robert Fitch.
\newblock Stochastic path planning for autonomous underwater gliders with
  safety constraints.
\newblock In {\em Proc. of IEEE/RSJ IROS}, 2019.

\end{thebibliography}

\end{document}